\documentclass[journal]{IEEEtran}
\IEEEoverridecommandlockouts
\usepackage{makecell}
\usepackage{cite}
\usepackage{amsmath,amssymb,amsfonts}
\usepackage{algorithmic}
\usepackage{graphicx}
\usepackage{textcomp}
\usepackage{xcolor}
\usepackage{booktabs}

\ifCLASSINFOpdf
\else
\fi
\begin{document}

\title{HiPerformer: A High-Performance Global–Local Segmentation Model with Modular Hierarchical Fusion Strategy}

\author{Dayu Tan,
        Zhenpeng Xu,
        Yansen Su,
        Xin Peng,
        Chunhou Zheng,
        and Weimin Zhong
\thanks{This work was supported in part by the National Key Research and Development Program of China (2021YFE0102100), in part by National Natural Science Foundation of China (62303014, 62172002, 62322301). (\emph{Corresponding author: Yansen Su.})}
\thanks{Dayu Tan, Zhenpeng Xu, Yansen Su, and Chunhou Zheng are with the Key Laboratory of Intelligent Computing and Signal Processing, Ministry of Education, Anhui University, Hefei 230601, China (e-mail: suyansen@ahu.edu.cn).}
\thanks{Xin Peng and Weimin Zhong are with the Key Laboratory of Smart Manufacturing in Energy Chemical Process, Ministry of Education, East China University of Science and Technology, Shanghai 200237, China (e-mail: wmzhong@ecust.edu.cn).}}


\maketitle

\begin{abstract}
Both local details and global context are crucial in medical image segmentation,  and effectively integrating them is essential for achieving high accuracy. However, existing mainstream methods based on CNN-Transformer hybrid architectures typically employ simple feature fusion techniques such as serial stacking, endpoint concatenation, or pointwise addition, which struggle to address the inconsistencies between features and are prone to information conflict and loss. To address the aforementioned challenges, we innovatively propose HiPerformer. The encoder of HiPerformer employs a novel modular hierarchical architecture that dynamically fuses multi-source features in parallel, enabling layer‑wise deep integration of heterogeneous information. The modular hierarchical design not only retains the independent modeling capability of each branch in the encoder, but also ensures sufficient information transfer between layers, effectively avoiding the degradation of features and information loss that come with traditional stacking methods. Furthermore, we design a Local-Global Feature Fusion (LGFF) module to achieve precise and efficient integration of local details and global semantic information, effectively alleviating the feature inconsistency problem and resulting in a more comprehensive feature representation. To further enhance multi-scale feature representation capabilities and suppress noise interference, we also propose a Progressive Pyramid Aggregation (PPA) module to replace traditional skip connections. Experiments on eleven public datasets demonstrate that the proposed method outperforms existing segmentation techniques, demonstrating higher segmentation accuracy and robustness. The code is available at https://github.com/xzphappy/HiPerformer.
\end{abstract}


% \fi
\IEEEpeerreviewmaketitle

\section{Introduction}
\IEEEPARstart {P}{recise} medical image segmentation is essential for accurately assessing lesion extent. The detailed information it provides enables physicians to gain a comprehensive understanding of the patient's condition, thereby facilitating the development of more targeted and effective treatment plans and significantly improving therapeutic outcomes.  In multi-region and fine-grained medical images, the anatomical structures are complex, and morphological differences are subtle.   Different tissues or lesions are extremely similar in grayscale and texture features, and they are often adjacent to each other, resulting in blurred segmentation boundaries. Target structures such as tumors or lesions, which have small volumes with significantly fewer pixels compared to normal tissues, are prone to being overlooked or misclassified due to their low resolution and limited semantic information.  Furthermore, the background of medical images often contains substantial extraneous noise originating from scanning artifacts, device interference, or surrounding tissues, which can lead to mis-segmentation.  The aforementioned challenges make the segmentation of multi-region and fine-grained medical images particularly difficult.  Traditional machine learning methods \cite{otsu1975threshold,tan2023deep} still require manual feature selection, and segmentation performance depends heavily on how accurately those features are chosen, making it difficult to handle complex segmentation tasks that require rich semantic information. In recent years, deep learning techniques are widely applied to medical image segmentation, effectively reducing subjective errors introduced by human judgment, improving annotation accuracy and consistency, and markedly enhancing segmentation performance.

U-Net\cite{ronneberger2015u} adopts an encoder-decoder architecture with a U-shaped structure, significantly enhancing the accuracy of medical image segmentation, and has become a classic and widely used model in this field.  It is based on a pure CNN architecture, which efficiently captures the local details and texture information of images through the sliding window of convolutional kernels.  However, limited by CNN's receptive field, it struggles to model long-range dependencies. In contrast, the Transformer employs a self-attention mechanism that effectively captures long-range dependencies among pixels at a global scale. The global modeling capability has been integrated into image segmentation tasks, resulting in the emergence of Swin-UNet\cite{cao2022swin}. Built upon a pure Transformer-based framework, Swin-UNet utilizes a U-shaped architecture to incorporate global contextual information effectively, yet its capability to perceive local details remains limited.

To fully leverage the respective advantages of CNN and Transformer, recent studies explore the integration of both architectures for image segmentation, aiming to preserve both local details and global semantic context, thereby enhancing overall performance. However, most existing mainstream hybrid architectures adopt simple approaches such as serial stacking (e.g., TransUnet\cite{chen2021transunet}), concatenation at the end of network (e.g., FAT-Net\cite{wu2022fat}), or pointwise addition (e.g., MixFormer\cite{liu2024mixformer}) and therefore neglect inconsistencies between convolutional and transformer-derived feature representations. Such neglect hampers the ability to balance the contributions of each feature type and can introduce feature conflicts and redundancy. In addition, existing studies commonly lack mechanisms for cross-level interaction, resulting in a separation between local details and global semantics, information loss during feature propagation, and an inability to achieve comprehensive and efficient fusion.

To more effectively address the aforementioned problem, we propose HiPerformer, a novel U-shaped network architecture. The architecture employs a modular hierarchical design in the encoder and effectively fuses CNN and Transformer features via a Local and Global Feature Fusion module, mitigating feature conflicts and information loss. In the output stage, a fusion mechanism is employed to combine features from four different stages, fully retaining multi-scale information and enhancing spatial detail and edge representation capability. Additionally, we replace traditional skip connections with a Progressive Pyramid Aggregation module, effectively suppressing noise propagation. The main contributions of our research can be summarized as follows:

\begin{figure*}[htbp]
  \centering
  \includegraphics[width=1\linewidth]{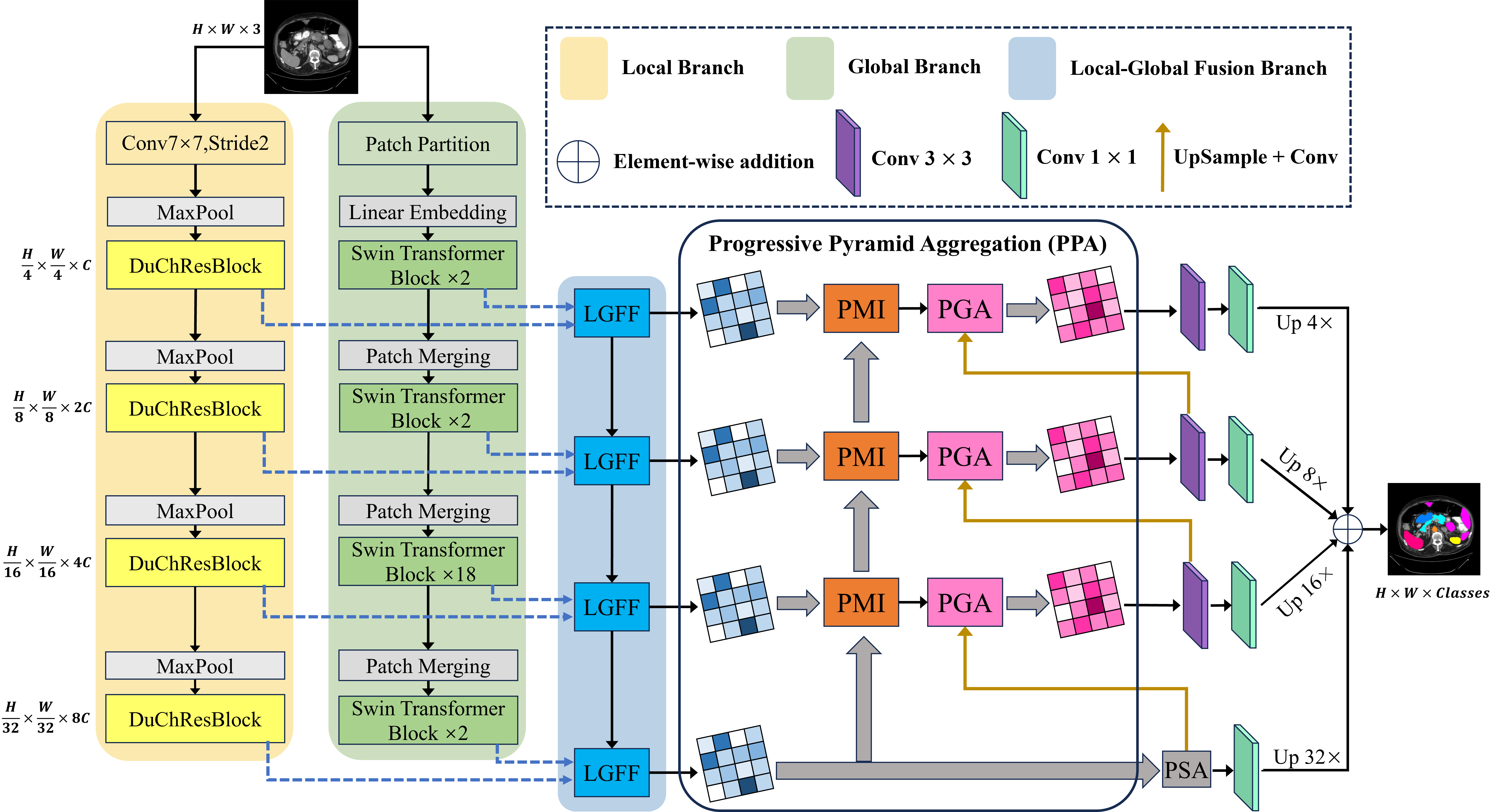}
  \caption{Illustration of the proposed HiPerformer. We utilize a modular hierarchical fusion strategy to redesign the encoder component, employ the Local and Global Feature Fusion (LGFF) module to efficiently and accurately merge local detail features with global semantic information, and adopt Progressive Pyramid Aggregation (PPA) to replace the traditional skip connections.}
  \label{fig}
\end{figure*}

\begin{itemize}
\item This study presents a novel modular hierarchical encoder that progressively integrates multi-source information layer by layer. The hierarchical fusion mechanism ensures effective information propagation across layers, preventing feature degradation associated with conventional stacked fusion.  Meanwhile, each branch of the encoder preserves its independent modeling capability, avoiding the loss of important information during fusion and preventing disruption of internal structures.  
\item We design a Local and Global Feature Fusion (LGFF) module that can efficiently and accurately fuse local detail features with global semantic information. The module mitigates feature inconsistency and achieves a more comprehensive and refined feature representation, thereby significantly improving the model's accuracy and robustness in complex scenarios. 
\item We propose a Progressive Pyramid Aggregation (PPA) module to replace traditional skip connections. The PPA module not only performs progressive multiplicative fusion of shallow and deep features to amplify feature differences and narrow the semantic gap, but also further enhances feature representation capability while suppressing interference from irrelevant regions, enabling the network to focus more on key regions. 
\end{itemize}

\section{Related Work}
\subsection{Hybrid CNN-Transformer Segmentation Network}
An increasing number of studies capture both local details and global contextual information simultaneously to improve model performance. For example, TransUNet\cite{chen2021transunet} combines convolutional neural networks and Transformers for medical image segmentation, significantly improving segmentation accuracy. DA-TransUnet\cite{sun2024transunet} further introduces dual attention into TransUNet, greatly enhancing feature extraction capability. However, the aforementioned models are constructed by serial stacking, and a main limitation is that each layer models only one type of dependency (global or local), causing global information to be lost during local modeling and making features prone to being overwhelmed. Moreover, such stacked architectures require very deep networks, which readily lead to feature attenuation.

FAT-Net\cite{wu2022fat} and MixFormer\cite{liu2024mixformer} adopt parallel architectures to optimize feature fusion strategies; however, such methods typically perform fusion using only simple end-stage concatenation or point-wise addition, which does not adequately account for inconsistencies between features and easily induces feature conflict and redundancy. The Performer\cite{tan2024performer} model introduces an innovative interaction mechanism designed for CNN and Transformer architectures to mutually enhance image feature extraction capabilities. TransFuse\cite{zhang2021transfuse} proposes a novel fusion technique whose core component, the BiFusion module, effectively integrates multilevel features from the CNN and Transformer branches, enabling capture of global information without relying on extremely deep networks while maintaining high sensitivity to low-level details. To further mitigate inter-feature inconsistency, it is necessary to continue exploring new hybrid CNN-Transformer segmentation architectures.

\subsection{Skip connections in U-shaped architectures}
Skip connections improve the recovery of spatial detail by introducing high-resolution features from the encoder into the decoder; they also help alleviate vanishing gradients and stabilize model training. In recent years, many improvements to skip connections are proposed. Perspective+\cite{hu2024perspective+} designs a Spatial Cross-Scale Integrator (SCSI) in its skip connections to enable coherent fusion of information across stages and better preserve fine-grained details.  FSCA-Net\cite{tan2024novel} employs a Parallel Attention Transformer (PAT) to strengthen spatial and channel feature extraction within skip connections, further reducing information loss caused by downsampling. SUet\cite{li2023sunet} proposes an efficient feature fusion (EFF) module based on multi-attention to achieve better fusion between skip connections and decoder features in U-shaped network. 

UNet++\cite{zhou2018unet++} addresses the semantic gap between encoder and decoder features by introducing nested, densely connected skip pathways and deep supervision, thereby producing more compatible fused features. DenseUNet\cite{cai2020dense} significantly enhances feature reuse, gradient propagation, and parameter efficiency by incorporating Dense Blocks and dense connectivity into both the encoder and decoder paths of U-Net. The effectiveness of traditional skip connections is largely constrained by the quality of the encoder’s extracted features: when the encoder’s representational capacity is limited, substantial irrelevant background noise can be passed directly to the decoder through the skip connections, degrading the image reconstruction process. In addition, discrepancies in feature distributions between the encoder and decoder exacerbate this semantic gap. Therefore, optimizing and redesigning skip-connection mechanisms to improve feature fusion, suppress noise transmission, and achieve more precise semantic alignment between the encoder and the decoder constitutes an important direction for our future research.

\subsection{Spatial and Channel attention}
Spatial attention\cite{fu2019dual} mechanisms enhance the representation of important regions by focusing on key areas in an image and dynamically adjusting the feature weights at different spatial locations. Channel attention mechanisms concentrate on the interrelationships among feature channels, typically employing squeeze-and-excitation operations to learn the importance weights of individual channels and thereby amplify responses of effective channels. For example, the core idea of the SE\cite{hu2018squeeze} attention mechanism is to adaptively assign distinct weights to each channel to strengthen useful features and suppress irrelevant information.

CBAM\cite{woo2018cbam} integrates both channel and spatial attention mechanisms and cleverly fuses average pooling and max pooling to aggregate information. However, two important drawbacks remain. First, it fails to capture spatial information at multiple scales to enrich the feature space. Second, the spatial attention focuses only on local regions and cannot establish long-range dependencies. In contrast, the PSA\cite{zhang2021epsanet} module can handle spatial information of multi-scale input feature maps and effectively establish long-range dependencies among multi-scale channel attention. The PSA module  module directly computes attention weights along the spatial dimension, typically via point-wise operations, and emphasizes capturing fine-grained spatial information with a lightweight structure and high computational efficiency. Nevertheless, during multi-scale feature fusion, PSA lacks adaptive modeling of semantic associations across different scales. Therefore, it is necessary to further optimize its attention mechanism to enhance feature discriminability and semantic awareness, thereby improving the overall representational performance of the model.

\section{Proposed Method}

\subsection{Overall Architecture }
The overall network architecture of the proposed HiPerformer  is illustrated in Fig. 1. To fully integrate local and global information while minimizing feature conflicts, the encoder adopts a modular hierarchical design with three branches, each retaining its own independent modeling capability. The Local branch is dedicated to extracting local detailed information. Specifically, it first applies a convolutional layer with a 7 ${\times}$ 7 kernel and a stride of 2, reducing the input image to half its original size. It then performs four stages of local feature extraction, each consisting of a max-pooling layer followed by a DuChResBlock. The DuChResBlock employs a dual-channel architecture that integrates standard convolution with dilated convolution and incorporates a residual connection, as shown in the left-hand side of Fig. 2. Such a design preserves local feature extraction capabilities while effectively expanding the receptive field, thereby enhancing the model’s ability to capture fine-grained details. The specific computation process of DuChResBlock can be expressed as follows:
\begin{equation}
x_{s+1}^{t}=x_{s}+f_{s}\left(f_{s}\left(x_{s}\right)\right),
\end{equation}\begin{equation}
x_{s+1}^{d}=x_{s}+f_{s}^{k}\left(f_{s}^{k}\left(x_{s}\right)\right),
\end{equation}
\begin{equation}
x_{s+1} = {Conv}({Concat}[x_{s+1}^t, x_{s+1}^d]), 
\end{equation}
in Eqs. (1), (2), and (3), $x_{s+1}^{d}$, $x_{s+1}^{d}$, and $x_{s+1}$ denote the features captured by standard convolution, dilated convolution, and the resulting merged features in stage $s+1$, respectively. $f_{s}$ and $f_{s}^{k}$  denote the standard convolution and dilated convolution with dilation rate $k$ in stage $s$. The symbol ${Conv}$ denotes a 1 ${\times}$ 1 convolution operation. 

\begin{figure}[htbp]
\centerline{\includegraphics[width=1\linewidth]{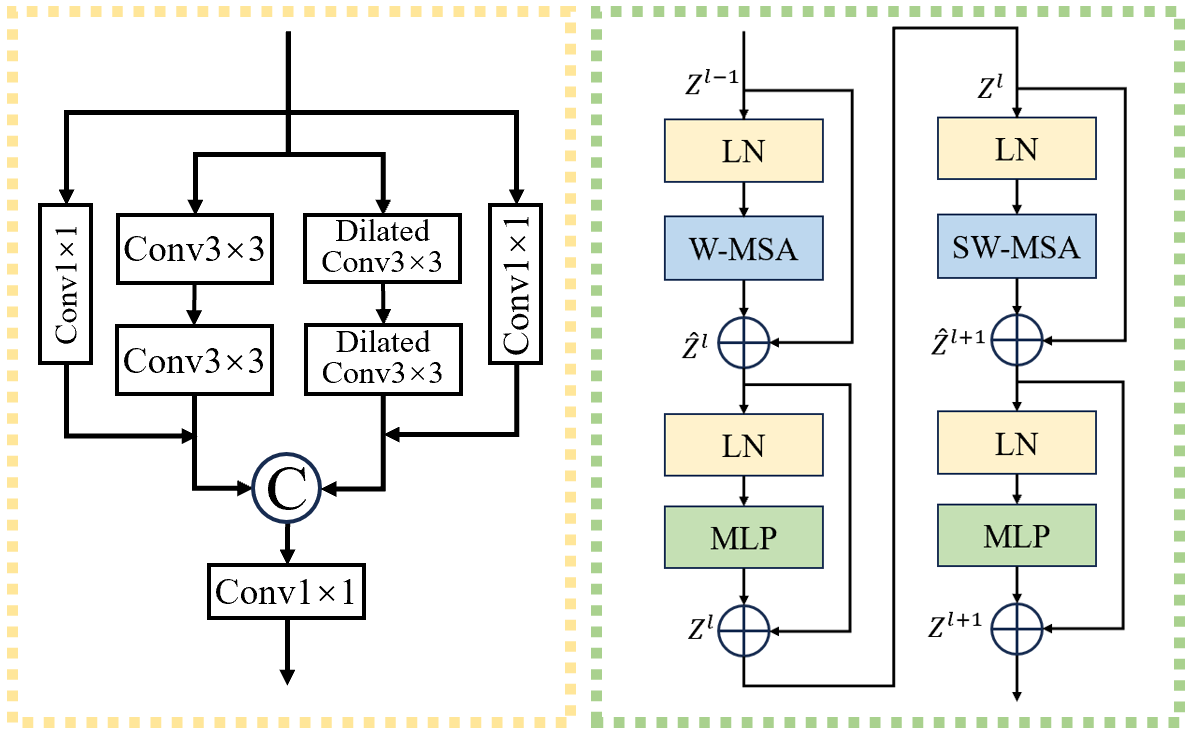}}
\caption{Feature extraction module. On the left is the fine-grained feature extraction module, and on the right is the coarse-grained feature extraction module.}
\label{fig}
\end{figure}

The Global branch consists of four stages comprising 2, 2, 18, and 2 Swin Transformer blocks, respectively, to capture global contextual information. Unlike the traditional Multi-Head Self-Attention (MSA) module, the Swin Transformer block is constructed based on shifted windows, as shown in the right-hand side of Fig. 2.  $\hat{z}^l$ and $z^l$ represent the outputs of the window based multi-head self attention (W-MSA) and MLP at the i-th layer, respectively, and can be expressed by the following formulas: 
\begin{equation}
    \hat{z}^l = \mathrm{W\text{-}MSA\bigl(LN(z^{l-1})\bigr)} + z^{l-1},
\end{equation}
\begin{equation}
    z^l = \mathrm{MLP}\bigl(\mathrm{LN}(\hat{z}^l)\bigr) + \hat{z}^l.
\end{equation}

The output after applying the shifted window partitioning strategy, through the shifted window-based
multi-head self attention (SW-MSA) and MLP modules, can be expressed by Eqs. (6) and (7), respectively.
\begin{equation}
    \hat{z}^{l+1} = \mathrm{SW\text{-}MSA\bigl(LN(z^l)\bigr)}+ z^l,
\end{equation}
\begin{equation}
    z^{l+1} = \mathrm{MLP}\bigl(\mathrm{LN}(\hat{z}^{l+1})\bigr) + \hat{z}^{l+1}.
\end{equation}

Each stage of the Local-Global fusion branch consists of an LGFF module that integrates both local and global features from the current layer with the fused features from the previous layer, thereby enabling multi-level feature aggregation.

In the bridge layer, we propose a novel PPA module. The PPA module integrates multi-scale features to mitigate the semantic gap while enhancing feature representation capability and suppressing interference from irrelevant information. To enhance detail representation and avoid boundary blurring, we additionally employ a multi-scale fusion strategy at the final stage of the decoder. Specifically, we first adjust the channel dimensions of outputs from four scales using 1 ${\times}$ 1 convolutions, then upsample them to a unified resolution, and finally fuse them through element-wise addition.

\subsection{LGFF: Local and Global Feature Fusion }
We innovatively proposed the LGFF module, as shown in Fig. 3a. Here, $G_{i}$ denotes the feature matrix output by the Transformer global feature block, $L_{i}$ represents the feature matrix output by the CNN local feature block, $F_{i-1}$ signifies the output feature matrix from the previous stage's LGFF module, and $F_{i}$ indicates the feature matrix generated through fusion in the current stage. The LGFF module efficiently integrates local features captured by the Local branch, global dependencies obtained by the Global branch, and semantic information from preceding layers. By effectively resolving feature learning inconsistencies within the same stage while mitigating feature discrepancies, it achieves a more comprehensive feature representation. 

\begin{figure}[htbp]
\centerline{\includegraphics[width=1\linewidth]{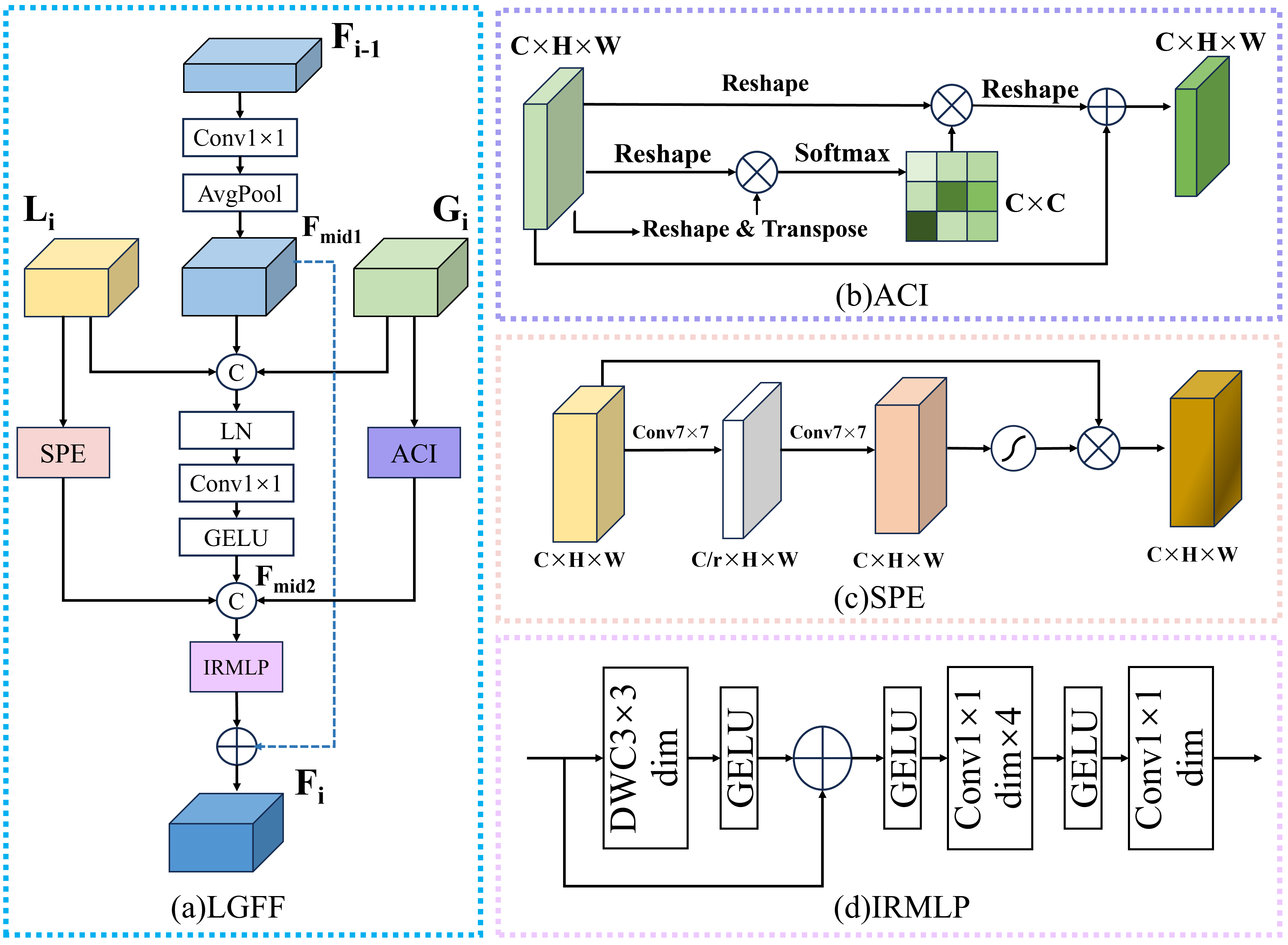}}
\caption{Structure diagram of the LGFF module and its submodules. (a) Local and global feature fusion (LGFF). (b) Adaptive channel interaction (ACI). (c) Spatial perception enhancement (SPE). (d) Inverted residual multilayer perceptron (IRMLP).}
\label{fig}
\end{figure}
Each channel map of high-level features contains category-specific semantic responses, which exhibit a certain degree of inter-channel correlation. By modeling the dependencies between channels, relevant feature maps can be reinforced, thereby enhancing the representation capability of specific semantics. To achieve that goal, we introduce a Adaptive channel interaction (ACI) for processing global features, as illustrated in Fig. 3b. The ACI module dynamically models inter-channel dependencies, adjusts the distribution of channel weights, and strengthens features related to the target semantics, ultimately improving the discriminative power of the global semantic representation. The formula of ACI can be expressed as follows:
\begin{equation}
ACI(x) = x + R\left( {Softmax}\left(R(x) \cdot R\&T(x)\right) \cdot R(x) \right),
\end{equation}
where ${Softmax}$ is the activation function, $R$ denotes the reshape operation, and  $T$ denotes the transpose operation.

We also introduced a spatial perception enhancement (SPE) to enhance the processing of local features, as illustrated in Fig. 3c. The SPE module employs two 7 ${\times}$ 7 convolutional layers to integrate spatial information and incorporates a reduction ratio $r$ to control the number of feature channels. By focusing on key areas and suppressing irrelevant background interference, it effectively enhances the representation of local details, thereby improving the ability to characterize fine-grained features. The formula of SPE can be expressed as follows:
\begin{equation}
SPE(x) = x \cdot {Sigmoid} \left( {Conv}_{7\times7} \left( {Conv}_{7\times7}(x) \right) \right),
\end{equation}
where $Sigmoid$ denotes the activation function, and ${Conv}_{7\times7}$ represents convolution operation with kernel sizes of 7 ${\times}$ 7.

The inverted residual multilayer perceptron (IRMLP) employs an inverted residual structure, combining high-dimensional representations with depthwise separable convolutions to effectively extract features with strong expressive power, as shown in Fig. 3d. The core of IRMLP in placing nonlinear operations in high-dimensional space, thus avoiding information loss caused by low-dimensional activation functions. The inverted residual structure not only enhances the efficiency of feature extraction but also effectively mitigates problems such as gradient vanishing and network degradation, facilitating stable training and performance improvement of deep networks. The formula of IRMLP can be expressed as follows:
\begin{equation}
\text{IRMLP(x)} = {Conv}_{1\times1}({Conv}_{1\times1}({DWConv}_{3\times3}(x) + x)),
\end{equation}
in Eq. (10), ${Conv}_{1\times1}$ denotes  convolution operation with kernel sizes of 1 ${\times}$ 1, and ${DWConv}_{3\times3}$ denotes  depthwise separable convolution operation with a kernel size of 3 ${\times}$ 3.

The overall process of LGFF can be expressed by the following formula: 
\begin{equation}
F_{mid1} = {Avgpool}(\text{Conv}_{1\times1}(F_{i-1})),
\end{equation}
\begin{equation}
F_{mid2} = {Conv}_{1\times1} ({Concat}[L_i, F_{mid1}, G_i]),
\end{equation}
\begin{equation}
F_{i} = \text{IRMLP}\left({Concat}\left[ACI(L_{i}), F_{mid2}, SPE(G_{i})\right]\right) + F_{mid1},
\end{equation}
herein, $Avgpool$ denotes the average pooling operation, and $F_{mid1}$ and $F_{mid2}$ denote the features generated at intermediate stages.

\subsection{PPA: Progressive Pyramid Aggregation}
We propose a novel Progressive Pyramid Aggregation (PPA) module to replace traditional skip connections in order to reduce the semantic gap and suppress noise interference, as shown in the middle part of Fig. 1. The PPA module is primarily composed of a Progressive Multiplicative Integration (PMI) module and a Pyramid Gated Attention (PGA) module.

1) Progressive Multiplicative Integration
(PMI): In U-Net architectures, features extracted by deep encoder layers are rich in high-level semantic information, whereas features from shallow encoder layers are better at capturing fine-grained boundary details. To bridge the semantic gap between them, existing methods typically fuse deep and shallow features. However, shallow encoder features often contain substantial background noise, causing most fusion strategies to be interfered by irrelevant information and thereby degrading final segmentation performance. To address the
aforementioned issue, we design the PMI module. The PMI module employs cascading multiplication to progressively fuse deep and shallow features through skip connections, enabling the aggregation of multi-scale semantic information. Specifically, deep features first undergo an upsampling process, and then both deep features and shallow features are simultaneously processed through a 3 ${\times}$ 3 convolutional layer and a 1 ${\times}$ 1 convolutional layer. The outputs are then fused using element-wise multiplication. Such multiplicative fusion amplifies the differences between noisy and normal regions, effectively suppressing background noise in shallow features and improving feature representation. The output  after processing the outputs $x_i$ ($i$=1, 2, 3, 4) of the four-level encoder through the PMI module is represented as follows:
\begin{equation}
    y_i = \begin{cases} 
f(x_i) \cdot f({Up}(y_{i+1})), & i = 1,2,3 \\
x_4, & i = 4 
\end{cases},
\end{equation}
in Eq. (14), $f(\cdot)$ represents 3 ${\times}$ 3 and 1 ${\times}$ 1 convolution used for channel adjustment and feature enhancement, and $Up$ denotes upsampling.

 % \subsection{PGA: Pyramid Gated Attention}
% We propose a novel PGA module (Fig. 4b), which is composed of two primary submodules: 
% 2) Pyramid Gated Attention (PGA): The PGA module (Fig. 4b) is composed of two primary submodules:
2) Pyramid Gated Attention (PGA): We propose a novel PGA module (Fig. 4a), which is composed of two primary submodules:
an EAG module and a PSA module. The EAG (Fig. 4b) module concatenates semantic features from the bridging layer and the decoder, doubling the feature dimensionality. A residual connection atop the AG mitigates the influence of high-level semantics on low-level semantics when input feature correlation is weak, thereby enhancing the semantic expressiveness of the concatenated features and effectively suppressing interference from irrelevant regions. The formula for EAG can be expressed as follows:
\begin{equation}
    EAG(e, d) = d + d \times \text{Sigmoid} \left( \text{Conv}_{1 \times 1} \left( \text{ReLU} (F_e + F_d) \right) \right),
\end{equation}
\begin{equation}
    F_{e} = {ReLU}\left(BN\left(Gp{Conv}_{1\times1}(e)\right)\right),
\end{equation}\begin{equation}
    F_{d} = {ReLU}\left(BN\left(Gp{Conv}_{1\times1}(d)\right)\right),
\end{equation}
herein, the $Sigmoid$ and $ReLU$ denote activation functions, and $BN$ represents batch normalization. ${Conv}_{1\times1} $ and $Gp{Conv}_{1\times1}$  refer to standard 1 ${\times}$ 1 convolution and group convolution respectively, while $e$ and $d$ represent features from the bridging layer and decoder.

\begin{figure}[htbp]
\centerline{\includegraphics[width=1\linewidth]{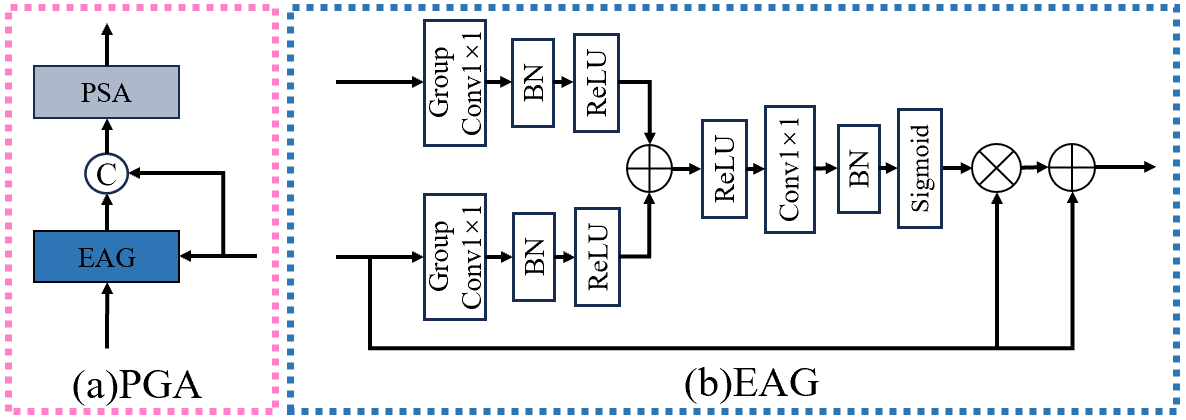}}
\caption{(a) Structure diagrams of PGA. (b) Structure diagrams of EAG.}
\label{fig}
\end{figure}

If the concatenated features are directly processed afterwards, it may result in a significant loss of useful information. To further emphasize key features and extract richer feature information, we feed the feature information obtained from EAG into the PSA module to more effectively guide the network in focusing on relevant areas, 0achieving accurate localization of lesions.

\begin{figure*}[htbp]
\centerline{\includegraphics[width=1\linewidth]{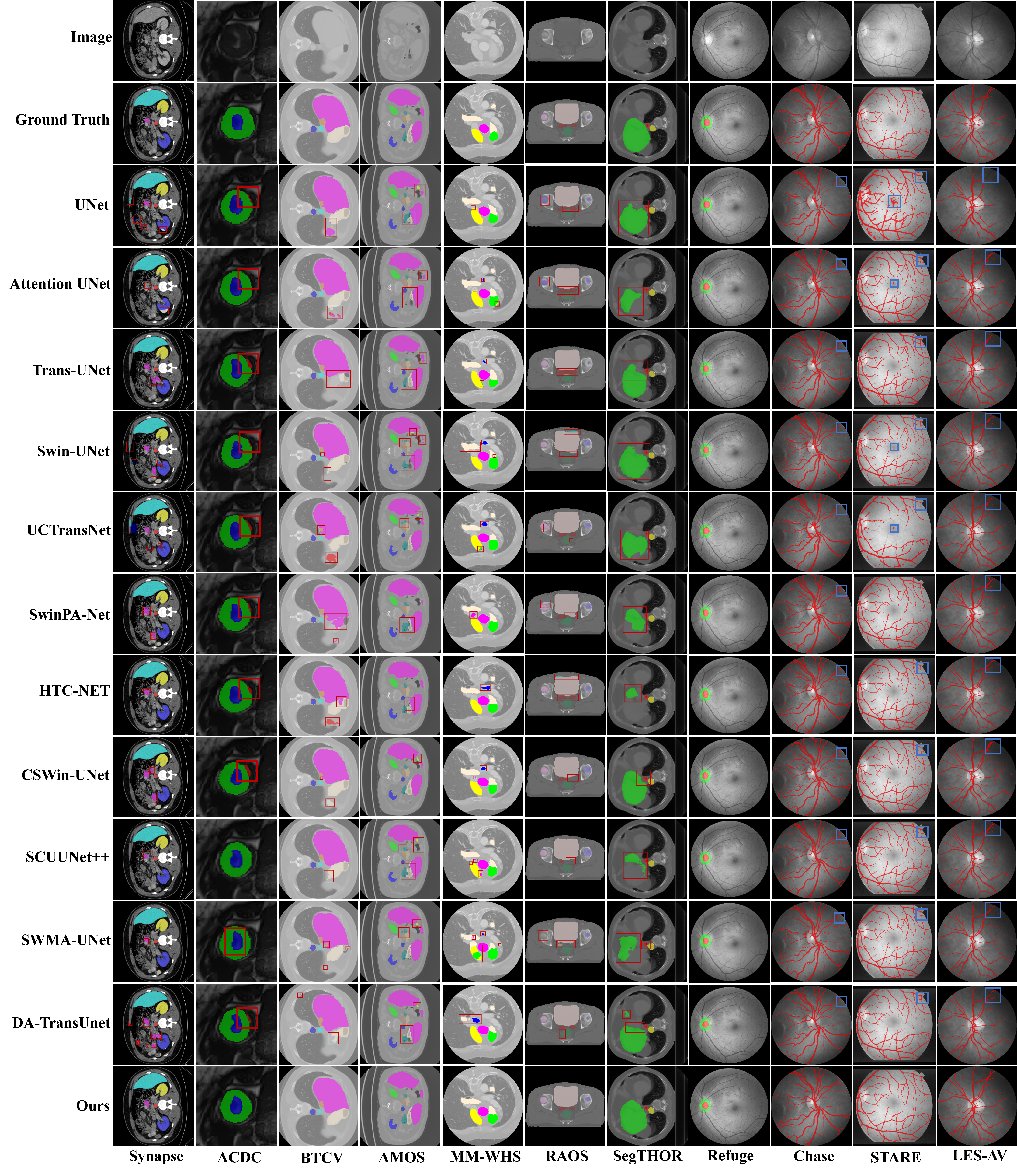}}
\caption{The visualization results of HiPerformer and eleven comparison methods for image segmentation on eleven datasets, as well as the original image and
GroundTruth. Incorrect segmentation areas are marked with red and blue boxes.}
\label{fig}
\end{figure*}

\subsection{Loss Function}
We use a weighted sum of the cross-entropy loss and the Dice loss as the loss function to balance pixel-level accuracy and global region consistency. The formula is as follows:
\begin{equation}
    L = \alpha L_{CE} + (1 - \alpha) L_{Dice},
\end{equation}
the hyperparameter $\alpha$ controls the relative weighting between $L_{CE}$ and $L_{Dice}$ in the final loss function. The formula for $L_{Dice}$ is presented in Eq. (19):
\begin{equation}
    L_{Dice} = 1 - \frac{2|X \cap Y|}{|X| + |Y|},
\end{equation}
where $X$ and $Y$ represent the cardinalities of the ground truth and predicted values, respectively. 

The formula for $L_{CE}$ is presented in Eq. (20): 
\begin{equation}
    L_{CE} = -\frac{1}{N} \sum_{i=0}^{N} \sum_{c=0}^{C} y_{i,c} \log(\hat{y}_{i,c}),
\end{equation}
herein, $N$ and $C$ represent the total number of samples and classes, respectively. $y_{i,c}$ is an indicator variable that equals 1 when sample $i$ belongs to class \textit{c}, and 0 otherwise. $\hat{y}_{i,c}$ represents the predicted probability of sample $i$ belonging to class $c$. 

\subsection{Evaluation Metrics}
To assess the performance of our approach across all datasets, we primarily employ the mean Dice Similarity Coefficient (DSC) and the mean Hausdorff distance at the 95th percentile (HD95) as our evaluation metrics. The formulas are as follows:
\begin{equation}
    DSC = \frac{2|X \cap Y|}{|X| + |Y|},
\end{equation}
\begin{equation}
    H(X, Y) = \max \left( \max_{x \in X} \min_{y \in Y} d(x, y), \max_{y \in Y} \min_{x \in X} d(x, y) \right),
\end{equation}
where $X$ represents the ground truth and $Y$ represents the predicted values, $ d(x, y)$ denotes the distance between pixel point $x$ and pixel point $y$.

For the Chase, STARE, and LES-AV datasets, we also add two evaluation metrics: average Recall and mean IOU.  The formulas are shown below:
\begin{equation}
\text{Recall} = \frac{\text{TP}}{\text{FN} + \text{TP}},
\end{equation}
\begin{equation}
    \mathrm{IoU} = \frac{\text{TP}}{\text{FP} + \text{FN} + \text{TP}},
\end{equation}
herein, true positive (TP) denotes the number of pixels correctly segmented as lesions, false
 negative (FN) is the number of true lesions pixels that are not segmented as normal tissue, and false positive (FP) denotes the number of background pixels wrongly segmented as lesions.

\section{  Experiments}
\subsection{Datasets}
We conduct experiments on eleven datasets, primarily focusing on multi-region and fine-grained segmentation. Below is a brief description of each dataset: 

\textbf{1) Synapse:} The dataset is derived from the MICCAI 2015 Multi-Atlas Abdomen Labeling Challenge\cite{landman2015miccai}. It includes abdominal CT scans from 30 patients, comprising a total of 3779 axial clinical CT images covering eight abdominal organs. In this study, all input images are uniformly resized and standardized to a resolution of 224 ${\times}$ 224 pixels. Consistent with previous research\cite{chen2021transunet},  the dataset is randomly divided into 18 cases (2212 axial slices) for training and 12 cases for testing.

\textbf{2) ACDC:} The dataset is derived from the Automatic Cardiac Diagnosis Challenge\cite{bernard2018deep} and provides manually annotated images for three regions, with pixel sizes varying between 0.83 and 1.75 ${mm}^2$. In this study, all input images are uniformly standardized to a resolution of 224 ${\times}$ 224 pixels. The dataset is randomly divided into 80 training cases (1510 axial slices) and 20 testing cases.

\textbf{3) BTCV:} The dataset from the BTCV Challenge\cite{landman2015miccai} focuses on the segmentation of 13 abdominal organs. The in-plane resolution of these images varies from 0.54 ${\times}$ 0.54 ${mm}^2$ to 0.98 ${\times}$ 0.98 ${mm}^2$, with slice thickness ranging from 2.5 $mm$ to 5.0 $mm$. In this study, we select a dataset of 30 cases, with input images standardized to a resolution of 224 ${\times}$ 224 pixels. The dataset is randomly divided into 24 training cases (1740 axial slices) and 6 testing cases.

\textbf{4) AMOS:} The dataset originates from the 2022 Multi-Modality Abdominal Multi-Organ Segmentation Challenge\cite{ji2022amos}. We select Task 1—Abdominal Organ Segmentation (CT imaging only), where each case includes voxel-level annotations for 15 abdominal organs. In this study, we randomly select 50 cases and divide them into 40 training cases (5783 axial slices) and 10 test cases. All input images in the dataset are standardized to a resolution of 224 ${\times}$ 224 pixels.

\textbf{5) SegTHOR:} The dataset is from the ISBI 2019 Challenge on Segmentation of Thoracic Organs at Risk in CT Images\cite{lambert2020segthor}. It contains CT scans of 40 cases, with each scan sized at 512 ${\times}$ 512 pixels, and each case includes voxel-level annotations for four thoracic organs. In this study, all input images are uniformly standardized to a resolution of 224 ${\times}$ 224 pixels. The dataset is randomly divided into 32 training cases (5995 axial slices) and 8 test cases.

\textbf{6) RAOS:} The dataset\cite{luo2024rethinking} is an abdominal multi-organ segmentation collection that includes cases with organ absence or morphological abnormalities. It comprises Computed Tomography (CT) images from 413 abdominal tumor patients who had undergone surgery or radiotherapy/chemotherapy, with annotations for 17 (female) or 19 (male) abdominal organs. Each case includes voxel-wise annotations for 19 abdominal organs. In this study, we randomly select a subset of 40 cases from the dataset. Input images are standardized to a resolution of 224 ${\times}$ 224 pixels. The dataset is randomly divided into 32 cases (5171 axial slices) for training and 8 cases for testing.

\begin{table*}[h!]
\centering
\setlength{\abovecaptionskip}{2pt}
\caption{Comparsion Result On The Synapse Dataset}
\label{tab:ablation}

\begin{tabular}{c c c c c c c c c c c}
    \toprule
    Method& DSC {\color{red}$\uparrow$}& HD95 {\color{red}$\downarrow$}& Aorta& Gallbladder& Kidney(L)& Kidney(R)& Liver& Pancreas& Spleen&Stomach\\
     & (\%, mean)& (mm, mean)& \multicolumn{8}{c}{DSC {\color{red}$\uparrow$}}\\
    \midrule
    UNet & 76.83& 29.877& 87.44& 66.42& 78.98& 72.03& 92.41& 56.47& 86.56&74.32
\\
 Attention UNet
& 79.70& 31.870& \textbf{89.91}& 71.57& 82.39
& 73.04& 94.19& 59.39& 90.21& 76.93\\
    Trans-UNet & 79.32& 22.773& 87.69& 61.92& 84.93& 82.76& 94.30& 60.91& 88.95&73.10
\\

    Swin-UNet & 78.12& 19.558& 84.99& 64.96& 83.81& 80.66& 93.96& 54.26& 89.01&73.33
\\
 UCTransNet
& 79.02& 27.509& 87.91& 64.72& 84.14& 74.63& 93.54& 62.77& 89.95&74.52\\
    SwinPA-Net & 81.11& 25.268& 86.73& 70.84& 82.29& 80.67& 94.22& \textbf{68.47}& 88.56&77.12
\\
    HTC-NET & 81.75& 28.826& 88.98& 67.35& 85.21& 83.42& 94.82& 62.68& 91.45&80.10
\\
 CSWin-UNet& 82.23& 20.565& 87.84& 69.03& \textbf{87.88}& 81.75& 94.77& 64.63& 90.73&81.24\\
    SCUNet++& 80.73& 21.531& 87.26& 67.90& 86.22& 82.82& 93.90& 60.32& 89.46&77.95
\\
    SWMA-Unet & 79.26& 25.206& 87.14& 68.60& 84.50& 81.21& 94.19& 55.61& 87.66&75.19
\\
 DA-TransUnet& 79.76& 22.047& 87.92& 68.38& 85.46& 81.98& 94.28& 58.50& 86.44&75.14\\
    \textbf{Ours} & \textbf{83.93}& \textbf{11.582}& 88.39& \textbf{71.58}& 86.58& \textbf{85.11}& \textbf{95.20}& 68.21& \textbf{92.15}&\textbf{84.20}\\
    \bottomrule
  \end{tabular}
\end{table*}

\begin{table*}[h!]
\centering
\setlength{\abovecaptionskip}{2pt}
\caption{Comparsion Result On The BTCV Dataset}
\label{tab:ablation}
\setlength{\tabcolsep}{2pt}
\begin{tabular}{c c c c c c c c c c c c c c c c}
    \toprule
    Method& DSC {\color{red}$\uparrow$}& HD95 {\color{red}$\downarrow$}& Spleen & Kidney(R) & Kidney(L) & Gallbladder & Esophagus & Liver & Stomach &Aorta & IVC & HPV\&SV & Pancreas & RAG &LAG\\
     & (\%, mean)& (mm, mean)& \multicolumn{13}{c}{DSC {\color{red}$\uparrow$}}\\
    \midrule
    UNet & 71.74& 34.054& 82.93& 74.65& 83.69& 59.57& 71.41& 92.58& 77.47&82.35& 71.94& 65.31& 63.28& 51.28&56.05
\\
 Attention UNet
 & 73.55& 27.550 & 84.96& 85.53 & 85.91& 64.48& 70.82& 93.69& 76.92& 85.15& 75.25& 65.64& 58.60& \textbf{56.26} &55.91
\\
    Trans-UNet & 69.43& 16.227& 89.73& 86.29& 89.06& 58.78& 70.10& 93.78& 66.97&82.26& 69.61& 62.61& 50.97& 49.55&32.88
\\
    Swin-UNet & 69.71& 15.949& 89.12& 77.74& 81.76& 59.45& 68.23& 93.93& 81.23&81.70& 67.37& 54.40& 55.15& 50.82&45.36
\\
UCTransNet & 68.89& 32.216& 78.81& 76.91& 81.95& 50.88& 68.49& 92.30& 72.02& 84.87& 66.01& 62.57& 56.87& 49.90&54.03  
\\
    SwinPA-Net & 75.63& 14.388& 92.09& \textbf{87.69}& \textbf{89.86}& 70.04& 71.83& 94.70& 84.87&87.40& 74.30& 64.46& 62.88& 46.82&56.24
\\
    HTC-NET & 74.86& 18.334& \textbf{92.48}& 81.35& 84.63& 65.06& 71.95& 94.73& 83.71&86.78& 74.98& 64.87& 62.97& 50.35&59.25
\\
    CSWin-UNet& 73.43& 15.851& 90.69& 80.79& 85.10& 63.70& 64.90& \textbf{95.01}& \textbf{85.91}& 83.44& 71.97& 64.71& 58.35& 50.46& 59.59
\\
    SCUNet++& 72.65& 14.245& 92.48& 79.84& 84.63& 63.68& 69.89& 94.78& \textbf{84.30}&84.82& 67.41& 64.87& 58.39& 46.09&56.25
\\
    SWMA-Unet & 69.41& 18.221& 90.42& 79.88& 83.77& 54.59& 64.05& 93.67& 79.84&84.05& 67.24& 53.35& 56.66& 50.30&44.52
\\
DA-TransUnet & 67.79& 27.134& 84.25& 78.11& 75.58& 51.06& 65.96& 92.23& 76.34& 80.23& 69.11& 56.82& 51.03& 47.65& 50.95 \\
    \textbf{Ours} & \textbf{76.39}& \textbf{13.621}& 90.62& 84.28& 81.66& \textbf{70.93}& \textbf{73.39}& 94.72& 83.82&\textbf{88.27}& \textbf{75.37}& \textbf{68.46}& \textbf{66.40}& 51.75&\textbf{63.38}\\
    \bottomrule
  \end{tabular}
\end{table*}

\begin{table*}[h!]
\centering
\setlength{\abovecaptionskip}{2pt}
\caption{Comparsion Result On The AMOS Dataset}
\label{tab:ablation}
\setlength{\tabcolsep}{0.8pt}
\begin{tabular}{c c c c c c c c c c c c c c c c c c}
    \toprule
    Method& DSC {\color{red}$\uparrow$}& HD95 {\color{red}$\downarrow$}& Spleen & Kidney(R) & Kidney(L) & Gallbladder & Esophagus & Liver & Stomach &Aorta & IVC & Pancreas& RAG& LAG& Duodenum&Bladder&Pr/Ut
\\
     & (\%, mean)& (mm, mean)& \multicolumn{13}{c}{DSC {\color{red}$\uparrow$}}\\
    \midrule
    UNet & 60.06  & 20.156 & 82.76 & 84.54 & 86.21 & 49.05 & 62.03 & 91.79 & 66.78 & 84.97 & 65.85 & 56.53 & 49.67 & 49.25 & 51.70 & 58.72 & 51.08 \\
Attention-UNet & 64.98  & 26.569 & 81.88 & 77.12 & 76.11 & 46.71 & 69.03 & 90.66 & 69.24 & 82.68 & 66.09 & 57.56 & 50.39 & 49.07 & 50.90 & 56.54 & 50.79 \\
Trans-UNet & 71.63  & 12.246 & 90.17 & 92.22 & 91.26 & 62.33 & 65.45 & 94.33 & 73.37 & 83.89 & 75.83 & 64.70 & 52.58 & 53.68 & 58.54 & 59.31 & 56.76 \\
Swin-UNet & 64.54  & 14.245 & 88.06 & 88.97 & 85.76 & 50.77 & 53.57 & 93.00 & 69.33 & 75.77 & 66.53 & 58.43 & 40.34 & 42.77 & 54.39 & 56.62 & 43.75 \\
UCTransNet & 66.34  & 18.357 & 84.60 & 80.31 & 83.55 & 48.84 & 61.91 & 92.48 & 66.49 & 82.32 & 67.68 & 57.86 & 51.10 & 42.33 & 49.77 & 70.00 & 55.93 \\
SwinPA-Net & 70.73  & 13.600 & 89.43 & 88.24 & 87.42 & 56.69 & 63.40 & 93.82 & 74.55 & 81.73 & 72.30 & 62.93 & 47.17 & 49.21 & 57.97 & 70.78 & 65.31 \\
HTC-NET & 71.76  & 16.388 & 89.41 & 90.75 & 89.62 & 62.70 & 68.87 & 94.05 & 74.49 & \textbf{85.37} & \textbf{78.59} & 65.04 & \textbf{53.24} & 45.57 & 58.41 & 60.90 & 59.31 \\
CSWin-UNet & 71.86  & 11.806 & 91.72 & 91.90 & \textbf{92.34} & 56.03 & 61.92 & 94.74 & 75.10 & 86.27 & 72.71 & 66.83 & 52.86 & 52.21 & 62.27 & 61.58 & 59.36 \\
SCUUNet++ & 69.53  & 17.524 & 88.08 & 87.75 & 87.93 & 54.78 & 61.35 & 93.27 & 75.36 & 83.73 & 71.09 & 57.14 & 43.70 & 39.67 & 54.86 & \textbf{73.66} & \textbf{70.54} \\
SWMA-UNet & 67.56 &  13.786 & 87.31 & 89.63 & 88.27 & 58.16 & 58.93 & 93.48 & 68.58 & 81.92 & 69.48 & 61.54 & 42.34 & 41.97 & 53.56 & 61.44 & 56.79 \\
DA-TransUnet & 71.73 & 13.927 & 92.91 & \textbf{92.32} & 92.22 & 60.57 & 66.23 & 93.45 & 72.96 & 85.70 & 76.15 & 66.44 & 52.96 & 55.66 & 57.94 & 57.14 & 53.27 \\
\textbf{Ours} & \textbf{75.33}  & \textbf{8.409} & \textbf{93.79} & 91.68 & 90.84 & \textbf{62.96} & \textbf{69.04} & \textbf{94.92} & \textbf{79.25} & \textbf{88.42} & 77.37 & \textbf{68.76} & 51.53 & \textbf{59.83} & \textbf{64.10} & 70.55 & 66.88 \\

    \bottomrule
  \end{tabular}
\end{table*}

\begin{table*}[h!]
\centering
\setlength{\abovecaptionskip}{2pt}
\setlength{\tabcolsep}{2pt}
\caption{Comparsion Result On The RAOS Dataset}
\label{tab:ablation}

\begin{tabular}{c c c c c c c c c c c c}
    \toprule
    Method& DSC {\color{red}$\uparrow$}& HD95 {\color{red}$\downarrow$}& Liver & Spleen & Kidney(L) & Kidney(R) & Stomach & Gallbladder & Esophagus & Pancreas & Duodenum\\
     & (\%, mean)& (mm, mean)& \multicolumn{9}{c}{DSC {\color{red}$\uparrow$}}\\
    \midrule
   UNet & 68.88 & 28.769 & 91.50 & 88.22 & 70.58 & 69.23 & 78.52 & 61.05 & 65.20 & 63.72 & 40.86 \\
Attention UNet & 73.09 & 26.847 & 92.34 & 93.18 & 81.45 & 72.61 & 81.07 & 62.78 & 68.33 & 66.91 & 50.80 \\
Trans-UNet & 75.28 & 16.721 & 93.36 & 93.33 & \textbf{86.49} & \textbf{85.65} & \textbf{82.04} & 56.00 & 71.34 & 69.86 & 49.45 \\
Swin-UNet & 71.08 & 20.551 & 94.55 & 91.41 & 82.90 & 84.07 & 78.16 & 55.24 & 68.03 & 62.79 & 47.23 \\
UCTransNet & 69.86 & 19.927 & 92.33 & 90.18 & 76.09 & 73.67 & 78.32 & 57.03 & 66.05 & 63.35 & 45.85 \\
SwinPA-Net & 74.09 & 14.099 & 94.24 & 92.18 & 86.27 & 83.29 & 81.48 & 52.18 & 71.91 & 66.47 & 51.96 \\
HTC-NET & 74.63 & 20.857 & 94.82 & 88.68 & 80.78 & 83.83 & 80.08 & 62.52 & 71.78 & 66.29 & 51.59 \\
CSWin-UNet & 76.38 & 15.204 & 93.30 & 91.43 & 83.50 & 76.14 & 81.07 & 62.09 & \textbf{75.38} & \textbf{70.52} & 52.83 \\
SCUUNet++ & 73.01 & 15.912 & 94.05 & 90.55 & 79.59 & 75.87 & 80.60 & 53.89 & 71.41 & 66.69 & 46.95 \\
SWMA-UNet & 70.62 & 19.051 & 93.71 & 91.32 & 79.06 & 76.36 & 79.00 & 48.88 & 69.04 & 62.79 & 41.50 \\
DA-TransUnet & 74.27 & 15.646 & 93.34 & 93.07 & 90.76 & 88.66 & 81.21 & 51.19 & 66.51 & 69.73 & 49.65 \\
\textbf{Ours} & \textbf{76.66} & \textbf{8.664} & \textbf{94.77} & \textbf{93.67} & 82.86 & 79.60 & 79.11 & \textbf{66.47} & 69.23 & 64.31 & \textbf{57.05} \\

\midrule
\midrule
Method & Colon & Intestine & Adrenal(L) & Adrenal(R) & Rectum & Bladder & HOF(L) & HOF(R) & Prostate & SV \\
& \multicolumn{11}{c}{DSC {\color{red}$\uparrow$}}\\
\midrule
UNet & 73.08 & 73.31 & 53.02 & 35.31 & 71.73 & 92.12 & 72.35 & 70.75 & 72.81 & 65.40 \\
Attention-UNet & 75.28 & 78.97 & 58.46 & 57.13 & 76.13 & 92.13 & 85.13 & 85.28 & 52.58 & 58.08 \\
Trans-UNet & 72.71 & 79.10 & 62.60 & 54.87 & 75.52 & \textbf{96.66} & 80.28 & 81.57 & 74.68 & \textbf{67.86} \\
Swin-UNet & 70.76 & 75.11 & 52.75 & 45.21 & 74.90 & 92.53 & 85.82 & 85.55 & 77.00 & 26.45 \\
UCTransNet & 71.74 & 74.98 & 56.06 & 55.31 & 74.42 & 92.30 & 83.71 & 83.47 & 56.41 & 35.93 \\
SwinPA-Net & 73.48 & 79.13 & 53.49 & 55.60 & 74.98 & 93.80 & 87.11 & 87.71 & 70.85 & 51.47 \\
HTC-NET & \textbf{75.32} & 77.50 & 62.62 & \textbf{64.95} & 77.52 & 91.12 & 89.64 & 86.47 & 62.24 & 53.27 \\
CSWin-UNet & 74.19 & 79.12 & 62.26 & 57.10 & 76.58 & 94.69 & 85.19 & 86.02 & \textbf{83.57} & 65.58 \\
SCUUNet++ & 72.14 & 74.95 & 58.12 & 55.72 & 75.47 & 93.01 & 86.66 & 88.92 & 82.79 & 39.71 \\
SWMA-UNet & 71.06 & 75.45 & 54.56 & 53.75 & 72.60 & 93.04 & 83.91 & 81.88 & 60.93 & 52.93 \\
DA-TransUnet & 72.25 & 77.53 & 56.38 & 58.70 & 76.38 & 93.58 & 90.13 & 91.15 & 72.65 & 38.28 \\
\textbf{Ours} & 73.75 & \textbf{80.56} & \textbf{65.14} & 61.71 & \textbf{79.55} & 95.26 & \textbf{93.43} & \textbf{94.52} & 62.98 & 62.55 \\
    \bottomrule
  \end{tabular}
\end{table*}

\begin{table*}[h!]
\centering
\setlength{\abovecaptionskip}{2pt}
\caption{Comparsion Result On The MM-WHS Dataset}
\label{tab:ablation}

\begin{tabular}{c c c c c c c c c c}
    \toprule
    Method& DSC {\color{red}$\uparrow$}& HD95 {\color{red}$\downarrow$}& LVM& LABC& LVBC& RABC& RVBV& AATH& PA\\
     & (\%, mean)& (mm, mean)& \multicolumn{7}{c}{DSC {\color{red}$\uparrow$}}\\
    \midrule
    UNet & 83.20& 21.035& 88.92& 81.39& 87.31& 86.23& 84.33& 77.93& 76.31
\\
Attention UNet  & 82.43 & 20.494 & 87.40 & 80.28 & 84.76 & 86.44 & 84.50 & 77.70 & 75.95 \\
    Trans-UNet & 81.34& 17.225& 85.10& 76.76& 85.15& 85.64& 82.93& 79.03& 74.76
\\
    Swin-UNet & 80.44& 20.220& 88.06& 79.45& 80.23& 83.40& 82.45& 81.12& 68.34
\\
UCTransNet     & 83.09 & 21.582 & 87.67 & 80.48 & 86.07 & 86.09 & 84.00 & 80.22 & 77.09 \\
    SwinPA-Net & 85.61& 11.289& 90.24& 83.36& 89.03& 88.23& \textbf{86.86}& 83.83& 77.74
\\
    HTC-NET & 84.20& 13.327& 88.41& 80.77& 87.69& 87.26& 83.75& 84.49& 77.01
\\
CSWin-UNet    & 85.54 & 16.273 & 82.79 & 82.58 & 88.88 & 88.34 & 86.18 & 83.66 & 79.34 \\
    SCUNet++& 84.29& 14.882& 89.45& 82.12& 88.32& 86.32& 85.36& 81.15& 77.33
\\
    SWMA-Unet & 80.70& 25.691& 86.96& 78.81& 84.23& 82.65& 82.42& 80.38& 69.45
\\
DA-TransUnet & 81.11 & 19.668 & 84.19 & 79.39 & 85.93 & 85.63 & 80.77 & 79.04 & 72.79 \\
    \textbf{Ours} & \textbf{86.52}& \textbf{10.654}& \textbf{90.54}& \textbf{84.37}& \textbf{90.03}& \textbf{88.38}& 86.45& \textbf{86.02}& \textbf{79.83}\\
    \bottomrule
  \end{tabular}
\end{table*}

\textbf{7) MM-WHS:} The dataset is provided by the Multi-Modality Whole Heart Challenge 2017\cite{zhuang2019evaluation}. Each sample comprises seven substructures of the heart. The in-plane average resolution of the images is 0.44 ${\times}$ 0.44 ${mm}^2$,
with an average slice thickness of 0.60 $mm$. In this study, we select 40 cases as the dataset. All input images are uniformly standardized to a resolution of 224 ${\times}$ 224 pixels. The dataset is randomly split into 32 training cases (5542 axial slices) and 8 testing cases.

\textbf{8) Refuge:} The dataset\cite{orlando2020refuge} originates from the MICCAI 2018 Retinal Fundus Glaucoma Challenge, with the main task being optic disc and cup segmentation. It contains 1200 fundus images. In this study, all input images are uniformly standardized to a resolution of 224 ${\times}$ 224 pixels. The dataset is randomly split into 960 cases for training and 240 cases for testing.

\textbf{9) Chase:} The dataset\cite{owen2011retinal} contains a total of 28 retinal fundus images from 28 subjects, each with a resolution of 999 ${\times}$ 960 pixels. All images are accompanied by manually annotated vessel segmentation labels created by professional medical personnel. In this study, all input images are uniformly standardized to a resolution of 384 ${\times}$ 384 pixels. The dataset is randomly split into 22 cases for training and 6 cases for testing.

\textbf{10) STARE:} The dataset is a publicly available retinal vessel segmentation dataset, consisting of 20 fundus images with a resolution of 700 ${\times}$ 605 pixels. The images cover various types of lesions, including macular degeneration, hypertensive retinopathy, and diabetic retinopathy. Each image is accompanied by professionally hand-annotated vessel segmentation maps. In this study, all input images are uniformly standardized to a resolution of 384 ${\times}$ 384 pixels. The dataset is randomly split into 16 cases for training and 4 cases for testing.

\textbf{11) LES-AV:} The dataset\cite{orlando2018towards} contains fundus images from 22 different patients. Among them, 21 images have a field of view (FOV) of 30 degrees with a resolution of 1444 ${\times}$ 1620 pixels, while one image has a FOV of 45 degrees and a resolution of 1958 ${\times}$ 2196 pixels. Each pixel corresponds to a physical size of 6 micrometers. In this study, all input images are uniformly standardized to a resolution of 384 ${\times}$ 384 pixels. The dataset is randomly split into 18 cases for training and 4 cases for testing.

\subsection{Implementation Details}
 We implement our HiPerformer model using PyTorch  on an NVIDIA RTX 3090 GPU with 24GB of memory. The model is trained using the AdamW optimizer with an initial learning rate of 1e-4 and a weight decay of 5e-4. A cosine annealing learning rate scheduler is employed, setting T\_max to 100 and eta\_min to 1e-6, over a total of 300 training iterations. To prevent gradient explosion and ensure stable training, gradient clipping is implemented. Furthermore, to mitigate overfitting, a series of data augmentation techniques are applied to the dataset prior to inputting images into the model, which include random flipping and rotations at multiple angles. For the Chase, STARE, and LES-AV datasets, the batch size is set to 4, while for the other datasets, the batch size is set to 16.

\begin{table}[h!]
\centering
\setlength{\abovecaptionskip}{2pt}
\caption{Comparsion Result On The ACDC Dataset}
\label{tab:ablation}

\begin{tabular}{c c c c c c }
    \toprule
    Method& DSC {\color{red}$\uparrow$}& HD95 {\color{red}$\downarrow$}& RV& Myo& LV\\
     & (\%, mean)& (mm, mean)& \multicolumn{3}{c}{DSC {\color{red}$\uparrow$}}\\
    \midrule
    UNet & 91.58& 1.236& 90.58& 89.66& 94.49
\\
Attention UNet  & 91.50 & 1.117 & 90.55 & 89.28 & 94.66 \\
    Trans-UNet & 90.88& 1.127& 89.27& 88.77& 94.62
\\
    Swin-UNet & 90.69& 1.397& 89.51& 88.67& 93.91
\\
UCTransNet      & 91.26 & 1.107 & 90.02 & 89.34 & 94.43 \\
    SwinPA-Net & 91.54& 1.103& 90.48& 89.51& 94.63
\\
    HTC-NET & 90.79& 1.195& 89.51& 88.89& 93.97
\\
CSWin-UNet      & 90.99 & 1.896 & 90.32 & 88.56 & 94.09 \\
    SCUNet++& 91.08& 1.097& 90.32& 88.88& 94.03
\\
    SWMA-Unet & 90.67& 1.169& 89.34& 88.47& 94.20
\\
DA-TransUnet    & 90.07 & 1.316 & 88.23 & 88.09 & 93.89 \\
    \textbf{Ours} & \textbf{91.98}& \textbf{1.071}& \textbf{90.98}& \textbf{90.14}& \textbf{94.83}\\
    \bottomrule
  \end{tabular}
\end{table}

\begin{table}[h!]
\centering
\setlength{\abovecaptionskip}{2pt}
\caption{Comparsion Result On The SegTHOR Dataset}
\label{tab:ablation}
\setlength{\tabcolsep}{2pt}
\begin{tabular}{c c c c c c c}
    \toprule
    Method& DSC {\color{red}$\uparrow$}& HD95 {\color{red}$\downarrow$}& Esophagus& Heart& Trachea& Aorta\\
     & (\%, mean)& (mm, mean)& \multicolumn{4}{c}{DSC {\color{red}$\uparrow$}}\\
    \midrule
    UNet & 82.47 & 5.889 & 63.30 & 91.80 & \textbf{86.68}& 88.10 \\
    Attention UNet  & 83.08 & 6.420 & 65.17 & 92.20 & 85.65 & 89.30 \\
    Trans-UNet & 82.91 & 5.357 & 64.71 & 90.01 & 86.25 & 88.67 \\
    Swin-UNet & 79.20 & 6.698 & 56.35 & 91.85 & 84.45 & 84.16 \\
    UCTransNet      & 81.38 & 5.982 & 60.91 & 91.59 & 86.20 & 86.83 \\
    SwinPA-Net & 82.90 & 6.278 & 64.88 & 92.62 & 85.44 & 88.25 \\
    HTC-NET & 83.09 & 5.825 & 66.70 & 92.10 & 85.30 & 88.27 \\
    CSWin-UNet      & 82.50 & 5.214 & 62.33 & 92.21 & 86.46 & 88.98 \\
    SCUNet++& 81.55 & 5.806 & 61.50 & 91.65 & 85.43 & 87.60 \\
    SWMA-Unet & 77.89 & 8.381 & 55.02 & 90.15 & 84.63 & 81.76 \\
    DA-TransUnet    & 81.60 & 7.146 & 63.58 & 90.74 & 86.00 & 86.07 \\
    \textbf{Ours} & \textbf{84.28} & \textbf{4.637} & \textbf{69.05} & \textbf{93.35} & 85.29& \textbf{89.45} \\
    \bottomrule
  \end{tabular}
\end{table}

\begin{table}[h!]
\centering
\setlength{\abovecaptionskip}{2pt}
\caption{Comparsion Result On The Refuge Dataset}
\label{tab:ablation}
\begin{tabular}{c c c c c}
    \toprule
    Method& DSC {\color{red}$\uparrow$}& HD95 {\color{red}$\downarrow$}& CUP& DISC\\
     & (\%, mean)& (mm, mean)& \multicolumn{2}{c}{DSC {\color{red}$\uparrow$}}\\
    \midrule
    UNet & 89.61 & 3.915 & 89.42 & 89.81 \\
Attention UNet & 89.54 & 4.047 & 89.43 & 89.66 \\
Trans-UNet & 88.80 & 4.218 & 88.22 & 89.39 \\
Swin-UNet & 88.82 & 4.590 & 88.59 & 89.05 \\
UCTransNet & 89.66 & 4.099 & 89.40 & 89.91 \\
SwinPA-Net & 89.83 & 3.781 & 89.47 & 90.19 \\
HTC-NET & 89.72 & 3.777 & 89.20 & 90.23 \\
CSWin-UNet & 89.03 & 4.003 & 88.58 & 89.49 \\
SCUUNet++ & 89.26 & 3.942 & 88.86 & 89.66 \\
SWMA-UNet & 88.60 & 4.457 & 88.21 & 88.98 \\
DA-TransUnet & 89.02 & 4.105 & 88.54 & 89.50 \\
\textbf{Ours} & \textbf{90.13} & \textbf{3.627} & \textbf{89.79} & \textbf{90.46} \\
    \bottomrule
  \end{tabular}
\end{table}

\begin{table}[h!]
\centering
\setlength{\abovecaptionskip}{2pt}
\caption{Comparsion Result On The Chase Dataset}
\label{tab:ablation}
\begin{tabular}{c c c c c}
    \toprule
    Method& DSC {\color{red}$\uparrow$}& HD95 {\color{red}$\downarrow$}& Recall {\color{red}$\uparrow$}& IoU {\color{red}$\uparrow$}\\
     & (\%, mean)& (mm, mean)& (\%, mean)& (\%, mean)\\
    \midrule
UNet          & 63.55 & 48.142 & 59.47 & 47.08 \\
Attention-UNet & 74.19 & 23.602 & 72.07 & 60.10 \\
Trans-UNet    & 76.08 & 19.515 & 73.72 & 62.00 \\
Swin-UNet     & 72.35 & 18.065 & 71.49 & 57.14 \\
UCTransNet    & 68.37 & 37.740 & 66.54 & 52.50 \\
SwinPA-Net    & 76.63 & 13.149 & 77.53 & 62.99 \\
HTC-NET       & 75.36 & 17.228 & 76.70 & 61.16 \\
CSWin-UNet   & 77.12 & 13.419 & 78.58 & 63.46 \\
SCUUNet++     & 75.39 & 14.845 & 74.17 & 60.59 \\
SWMA-UNet    & 74.13 & 17.278 & 72.67 & 59.41 \\
DA-TransUnet  & 75.62 & 20.937 & 74.73 & 61.13 \\
\textbf{Ours} & \textbf{77.70} & \textbf{12.792} & \textbf{78.71} & \textbf{64.40} \\
    \bottomrule
  \end{tabular}
\end{table}

\begin{table}[h!]
\centering
\setlength{\abovecaptionskip}{2pt}
\caption{Comparsion Result On The STARE Dataset}
\label{tab:ablation}
\begin{tabular}{c c c c c}
    \toprule
    Method& DSC {\color{red}$\uparrow$}& HD95 {\color{red}$\downarrow$}& Recall {\color{red}$\uparrow$}& IoU {\color{red}$\uparrow$}\\
     & (\%, mean)& (mm, mean)& (\%, mean)& (\%, mean)\\
    \midrule
UNet & 51.58 & 31.818 & 61.74 & 40.24 \\
Attention UNet & 57.64 & 28.331 & 55.89 & 44.50 \\
Trans-UNet & 64.96 & 20.029 & 68.00 & 51.32 \\
Swin-UNet & 68.89 & 11.408 & 67.68 & 55.03 \\
UCTransNet & 48.96 & 51.213 & 47.07 & 37.20 \\
SwinPA-Net & 71.32 & 9.600 & 72.07 & 57.62 \\
HTC-NET & 69.13 & 13.446 & 68.42 & 55.57 \\
CSWin-UNet & 72.72 & 10.215 & 72.50 & 59.60 \\
SCUUNet++ & 69.99 & 11.702 & 73.49 & 56.51 \\
SWMA-UNet & 68.95 & 12.707 & 68.78 & 55.43 \\
DA-TransUnet & 67.96 & 16.726 & 66.79 & 54.42 \\
\textbf{Ours} & \textbf{73.26} & \textbf{8.405} & \textbf{75.41} & \textbf{60.29} \\
    \bottomrule
  \end{tabular}
\end{table}

\begin{table}[h!]
\centering
\setlength{\abovecaptionskip}{2pt}
\caption{Comparsion Result On The LES-AV Dataset}
\label{tab:ablation}
\begin{tabular}{c c c c c}
    \toprule
    Method& DSC {\color{red}$\uparrow$}& HD95 {\color{red}$\downarrow$}& Recall {\color{red}$\uparrow$}& IoU {\color{red}$\uparrow$}\\
     & (\%, mean)& (mm, mean)& (\%, mean)& (\%, mean)\\
    \midrule
UNet & 62.30 & 150.187 & 55.91 & 41.86 \\
Attention UNet & 75.48 & 45.245 & 72.39 & 58.99 \\
Trans-UNet & 77.15 & 40.094 & 75.98 & 60.68 \\
Swin-UNet & 74.70 & 30.300 & 71.29 & 58.35 \\
UCTransNet & 66.03 & 96.168 & 60.66 & 47.03 \\
SwinPA-Net & 76.45 & 26.129 & 77.04 & 60.29 \\
HTC-NET & 74.82 & 30.577 & 73.64 & 58.36 \\
CSWin-UNet & 77.35 & 25.252 & 77.27 & 61.92 \\
SCUUNet++ & 75.99 & 26.943 & 74.15 & 59.70 \\
SWMA-UNet & 74.93 & 34.573 & 70.77 & 58.34 \\
DA-TransUnet & 77.12 & 46.382 & 73.95 & 60.19 \\
\textbf{Ours} & \textbf{77.55} & \textbf{23.120} & \textbf{79.76} & \textbf{62.36} \\
    \bottomrule
  \end{tabular}
\end{table}

\subsection{Comparison Experiments}
To demonstrate the effectiveness of our model in segmentation tasks, we evaluate the performance of HiPerformer by comparing it with seven state-of-the-art methods: UNet\cite{ronneberger2015u}, Attention UNet\cite{oktay2018attention}, Trans-UNet\cite{chen2021transunet}, UCTransNet\cite{wang2022uctransnet}, Swin-UNet\cite{cao2022swin}, SwinPA-Net\cite{du2022swinpa}, HTC-NET\cite{tang2024htc}, CSWin-UNet\cite{liu2025cswin}, SCUNet++\cite{chen2024scunet++}, and SWMA-UNet\cite{tang2024swma}, DA-TransUnet\cite{sun2024transunet}. For each dataset, we assess the DSC for each organ as well as the average DSC and average HD95 across all organs. The best results are highlighted in bold.

1) Experiments on the Synapse dataset: Table I presents the comprehensive evaluation results of our method on the Synapse dataset. Compared to existing methods, our approach achieves superior performance in both the average DSC and HD95 metrics, reaching 83.93\% and 11.582, respectively. In the segmentation tasks of specific organs, our method attains the best performance on the Gallbladder, Right Kidney, Liver, Spleen, and Stomach.

The first column in Fig. 5 shows a qualitative comparison of HiPerformer with existing segmentation methods on the Synapse dataset. It can be observed that other existing methods often suffer from mis-segmentation of background areas or incomplete segmentation, especially for the Pancreas organ labeled in pink. In contrast, our method’s predictions are closer to the ground truth labels, demonstrating higher accuracy and consistency.

\begin{figure*}[htbp]
\centerline{\includegraphics[width=1\linewidth]{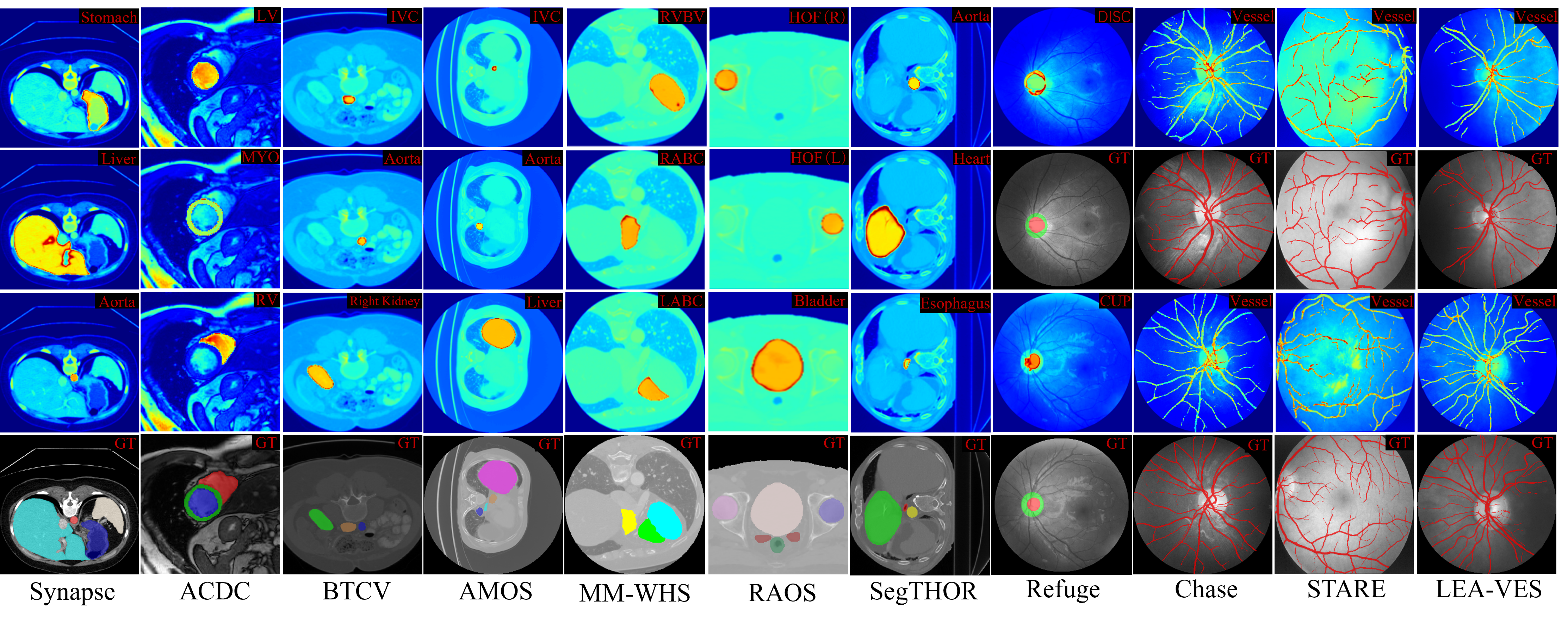}}
\caption{The visualization images of attention weight heatmaps on multiple datasets.}
\label{fig}
\end{figure*}

2) Experiments on the ACDC dataset: Table VI presents a comprehensive evaluation on the ACDC dataset. According to the results, our method outperforms existing approaches across all metrics, achieving an average DSC of 91.98\% and an HD95 of 1.071.

The second column in Fig. 5 shows a qualitative comparison of HiPerformer with existing segmentation methods on the ACDC dataset. Incorrectly segmented regions are marked with red boxes. We observe that most existing methods mistakenly segment parts of the background as RV, whereas our proposed method does not exhibit this issue, resulting in more accurate segmentation.

3) Experiments on the BTCV dataset: Table II presents a comprehensive evaluation on the BTCV dataset. We report the results on all 13 organs along with the corresponding mean performance over all organs. According to the results, our model achieves an average DSC of 76.39\% and an average HD95 of 13.621, both outperforming existing methods. Notably, for the smaller organ LAG, our method still performs excellently, achieving a DSC of 63.38\%, which is 3.97\% higher than the second-best method, CSWin-UNet.

The third column in Fig. 5 shows a qualitative comparison of  HiPerformer with existing segmentation methods on the BTCV dataset. Incorrectly segmented regions are marked with red boxes. It can be observed that other existing methods exhibit inaccuracies during segmentation, whereas our method shows precise organ segmentation with improved boundary delineation.

4) Experiments on the AMOS dataset: Table III presents a comprehensive evaluation on the AMOS dataset. Compared with existing methods, our approach performs better on both the mean DSC and HD95 metrics, achieving 75.33\% and 8.409, respectively.

The fourth column in Fig. 5 shows a qualitative comparison between HiPerformer and existing segmentation methods on the AMOS dataset. Incorrectly segmented regions are marked with red boxes. It can be observed that existing segmentation methods often misclassify the background as the gallbladder (yellow label), resulting in significant deviations in the outcomes. Additionally, the duodenum (deep cyan label) is also prone to missegmentation, which affects the overall quality. In contrast, our proposed method effectively addresses these issues, providing more precise and stable segmentation results, and demonstrating its advantages in detail feature extraction and boundary information capture.
It can be observed that many existing segmentation methods often misclassify background as the gallbladder (yellow label), leading to large deviations in the results. At the same time, the duodenum (dark cyan label) is also prone to incorrect segmentation, which degrades overall quality. In contrast, our proposed method effectively addresses these issues, yielding more accurate and stable segmentation results and demonstrating superior ability to extract fine-grained features and capture boundary information.

5) Experiments on the MM-WHS dataset: Table V presents a comprehensive evaluation on the MM-WHS dataset. According to the results, except for the RVBV organ, our method outperforms existing methods in the segmentation performance of all other organs, achieving an average DSC and HD95 of 86.52\% and 10.654, respectively.

The fifth column in Fig. 5 shows a qualitative comparison between HiPerformer and existing segmentation methods on the MM-WHS dataset. Incorrectly segmented regions are marked with red boxes. It can be observed that other existing methods produce some small and inaccurate error pixels during segmentation, whereas our method’s segmentation results are closer to the ground truth. The improvement in segmentation performance likely stems from fully capturing and integrating both local details and global context.

6) Experiments on the RAOS dataset: Table IV presents a comprehensive evaluation on the RAOS dataset. According to the results, among existing methods CSWin-UNet performs comparatively well, with an average DSC of 76.38\%. Building on this, our method achieves a further improvement, raising the average DSC to 76.66\%. Although the increase in average DSC is small, our method shows a clear advantage on the average HD95 metric, which is reduced by 6.54 and is significantly better than CSWin-UNet.

The sixth column in Fig. 5 shows a qualitative comparison between HiPerformer and existing segmentation methods on the RAOS dataset. Incorrectly segmented regions are marked with red boxes. It can be observed that existing methods often produce segmentation errors for the brown-labeled seminal vesicles (SV) and the pink-labeled head of the right femur (HOF(R)), resulting in large deviations from the ground truth.  In contrast, our method shows a clear advantage in segmenting these two critical regions, delivering more accurate and stable results that better match the true anatomical structures.

7) Experiments on the SegTHOR dataset: Table VII presents a comprehensive evaluation on the SegTHOR dataset. According to the results, except for the Trachea organ, the proposed method outperforms existing methods in the segmentation performance of all other organs, achieving an average DSC of 84.28\% and an HD95 of 4.637.

The seventh column in Fig. 5 shows a qualitative comparison of HiPerformer with existing segmentation methods on the SegTHOR dataset. Incorrectly segmented regions are marked with red boxes. We observe that existing methods exhibit missing regions in the segmentation of the heart, resulting in incomplete segmentation, whereas the proposed method is able to achieve complete segmentation of the heart.

8) Experiments on the Refuge dataset: Table VIII presents a comprehensive evaluation on the Refuge dataset. According to the results, the proposed method outperforms existing segmentation methods both in terms of performance on specific organs and overall average performance, achieving an average DSC of 90.13\% and an average HD95 of 3.627. 

The eighth column in Fig. 5 provides a qualitative comparison between HiPerformer and existing segmentation methods on the Refuge dataset. It can be observed that the proposed method shows no significant difference from existing methods in the visualization of segmentation results. The above results can largely be attributed to the limited number of organ classes in the dataset and the distinct feature differences between categories, which reduced the complexity of the segmentation task and resulted in high accuracy across all models. However, it is worth highlighting that our method offers a distinct advantage in accurately capturing organ boundary information. Enhanced representation of boundary details directly yields superior quantitative evaluation results.

9) Experiments on the Chase dataset: Table IX presents a comprehensive evaluation on the Chase dataset. According to the results, the proposed method performs excellently across all metrics, with an average DSC of 77.70\%, an average HD95 of 12.792, an average Recall of 78.71\%, and an average IoU of 64.40\%, all significantly outperforming existing methods.

The ninth column in Fig. 5 shows a qualitative comparison of HiPerformer with existing segmentation methods on the Chase dataset. Incorrectly segmented regions are marked with blue boxes. Due to uneven background illumination in the Chase images, low contrast of the blood vessels, and the relatively broad morphology of small arteries, the segmentation of retinal blood vessels is particularly challenging.  As shown in the area marked by the blue box, the blood vessels are thin and have poor contrast, making it difficult for existing methods to achieve complete and accurate segmentation, often resulting in breaks or omissions.     In contrast, our proposed method employs a modular hierarchical fusion strategy, which effectively reduces feature information loss during transmission and significantly improves the recovery of tiny capillaries, making the segmentation results closer to the ground truth.

10) Experiments on the STARE dataset: Table X presents a comprehensive evaluation on the STARE dataset. According to the results, our method outperforms existing methods across all evaluation metrics, achieving an average DSC of 73.26\%, an average HD95 of 8.405, an average Recall of 75.41\%, and an average IoU of 60.29\%.

The tenth column in Fig. 5 shows a qualitative comparison of HiPerformer with existing segmentation methods on the STARE dataset. Incorrectly segmented regions are marked with blue boxes. We select two typical regions prone to segmentation errors for analysis.  The first region is the gray area located in the center of the image, whose color is highly similar to blood vessels, causing models such as UNet, Attention UNet, Swin-UNet, and UCTransNet to mistakenly segment it as blood vessels.  The second region is the terminal structure of the blood vessels in the upper right corner.  Due to the low contrast between the pixels in this area and the background, the fine vessel terminals are difficult to completely separate from the background.  Except for SwinPA-Net, other existing methods fail to achieve complete segmentation here.  In contrast, our method introduces the Pyramid Gated Attention (PGA) module, which enhances feature representation of key regions while suppressing interference from irrelevant areas, thereby enabling more accurate differentiation of pathological regions. Furthermore, our approach efficiently integrates local details with global contextual information, allowing precise extraction of vessel terminal edge features and significantly improving segmentation accuracy.

11) Experiments on the LES-AV dataset: Table XI presents a comprehensive evaluation on the LES-AV dataset. According to the results, CSWin-UNet performs excellently, with an average DSC of 77.35\%. Our method, however, achieves even better performance, with the average DSC improved to 77.55\%, an increase of 0.2\%. In other metrics, the average HD95 is 23.120, the average Recall reaches 79.76\%, and the average IoU is 62.36\%, all of which outperform existing methods.

The eleventh column in Fig. 5 shows a qualitative comparison of HiPerformer with existing segmentation methods on the LES-AV dataset. Incorrectly segmented regions are marked with blue boxes. The retinal vascular network structure is extremely complex,  and the uneven distribution of pixel intensities leads to suboptimal performance of existing methods when handling crosses structures in dense vascular regions. As shown in the blue boxed area,  current methods often struggle to accurately capture edge features at vessel crossings,  which frequently results in vessel merging or partial vessel segmentation omissions. In contrast,  our method proposes the PMI module to achieve effective fusion of multi-scale semantic information significantly reducing boundary blurring.  our method enables more precise segmentation of intersecting vessels,  greatly enhancing the restoration and detail representation of the retinal vascular structure,  and fully demonstrating its superior performance in complex vascular environments.

\subsection{Loss parameter experiment}
To further optimize the model, we adopt a balanced joint loss function, which is a weighted combination of cross-entropy loss ($ L_{CE}$) and Dice loss ($ L_{Dice}$).  In Eq. (18), the value of the parameter a influences the model's optimization direction.  A larger $\alpha$ value causes the model to prioritize maintaining the overall consistency of the segmentation regions, potentially at the cost of pixel-level classification accuracy.  Conversely, a smaller $\alpha$ value leads to better performance in pixel-level classification but may hinder the optimization of overall segmentation consistency.  On the Synapse dataset, we systematically study the impact of a on segmentation accuracy by conducting experiments with $\alpha$ values of 0.1, 0.3, 0.5, 0.7, and 0.9.  The experimental results are shown in Fig. 7, indicating that when $\alpha$ = 0.5, the model achieves the best segmentation performance, with the highest average DSC of 83.93\% and the lowest average HD95 of 11.582.
\begin{figure}[htbp]
\centerline{\includegraphics[width=1\linewidth]{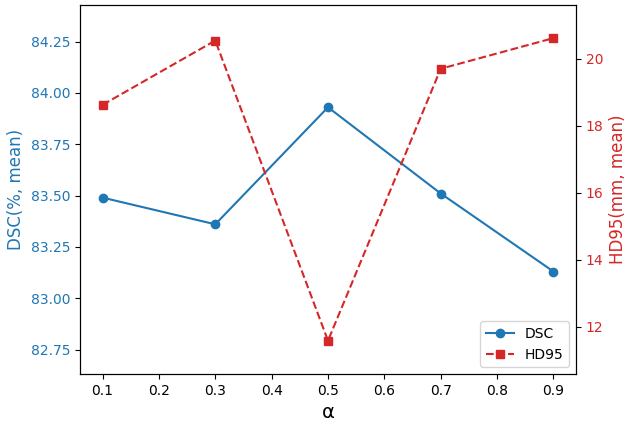}}
\caption{The experimental results of the combined loss function with different hyperparameters on the Synapse dataset.}
\label{fig}
\end{figure}

\subsection{Ablation Studies}
To evaluate the impact of each key component in the proposed method on the overall segmentation performance, a series of ablation experiments are conducted based on the Synapse and BTCV datasets. The results of the ablation studies are presented in Table XII. When the encoder retains only the Local branch, the average DSC on the Synapse and MM-WHS datasets decreases by 5.47\% and 7.62\%, respectively. When only the Global branch is retained, the average DSC drops by 1.70\% and 2.45\%, respectively. The above results fully validate the significant contribution of fusing CNN and Transformer to performance improvement.

Furthermore, when both the PMI and PGA modules in the skip connections are removed simultaneously, the average DSC on the Synapse and MM-WHS datasets decreases by 1.70\% and 1.71\%, respectively. Removing the PMI module alone leads to average DSC drops of 0.79\% and 1.44\%, respectively, while removing the PGA module alone causes average DSC declines of 0.94\% and 0.51\%, respectively. The above results indicate that the synergistic effect between the PMI and PGA modules is crucial, and combining them into a PPA module to replace traditional skip connections plays a decisive role in performance improvement.

When all modules are integrated, HiPerformer achieves the best segmentation performance. If any module is removed, performance decreases, which demonstrates the effectiveness and necessity of the designed module combination.

\begin{table}[h!]
\centering
\setlength{\abovecaptionskip}{2pt}
\caption{Results of Ablation Experiments on the Synapse and BTCV Datasets}
\label{tab:ablation}
\begin{tabular}{ccccc cc}\toprule

\multicolumn{5}{c}{Ablation Type} & \multicolumn{2}{c}{DSC {\color{red}$\uparrow$}(\%, mean)} \\
Local& Global& LGFF& PMI& PGA& Synapse& BTCV\\\midrule

$\checkmark$ & $\times$ & $\times$ & $\checkmark$ & $\checkmark$ & 78.46 & 68.77 \\
$\times$ & $\checkmark$ & $\times$ & $\checkmark$ & $\checkmark$ & 82.23 & 73.94 \\
$\checkmark$ & $\checkmark$ & $\checkmark$ & $\times$ & $\times$ & 82.83 & 74.68 \\
$\checkmark$ & $\checkmark$ & $\checkmark$ & $\times$ & $\checkmark$ & 83.14 & 74.95 \\
$\checkmark$ & $\checkmark$ & $\checkmark$ & $\checkmark$ & $\times$ & 82.99 & 75.88 \\
$\checkmark$ & $\checkmark$ & $\checkmark$ & $\checkmark$ & $\checkmark$ & \textbf{83.93} & \textbf{76.39} \\\bottomrule

\end{tabular}
\end{table}

\subsection{Attention Mechanism Visualization}
In medical image segmentation tasks, incorporating attention mechanisms not only improves the segmentation performance of the model but also enables intuitive visualization of the regions the model focuses on or suppresses during segmentation through attention heatmaps. As shown in Fig. 6, we present the segmentation heatmaps of partial organs across eleven datasets using the proposed method, where brighter colors indicate higher model attention. The results show that in the three retinal vessel segmentation datasets (Chase, STARE, and LEA-VES), the model tends to focus on the optic disc and cup regions, while struggling to effectively concentrate on the small and low-contrast vascular structures. The above-mentioned limitation is likely the main reason for its relatively lower accuracy in the retinal vessel segmentation task. In the ACDC dataset, the model also exhibits a small amount of erroneous attention to background regions. Apart from the datasets mentioned above, our model accurately focuses on target regions and effectively suppresses background noise, clearly demonstrating the superiority of our method.

\section{Conclusion}
In this study, we propose HiPerformer, a model specifically designed for multi-region and fine-grained medical image segmentation tasks.   Its encoder designs a novel modular hierarchical architecture and incorporates a Local and Global Feature Fusion module that hierarchically integrates local features with global context, effectively preventing information loss and feature conflicts from simple concatenation and enabling more accurate, efficient information integration.   At the skip connections, the model proposes a Progressive Pyramid Aggregation module, which not only fuse deep and shallow features but also employ targeted feature-enhancement mechanisms to effectively reduce semantic gaps across scales and suppress noise. We conduct a comprehensive evaluation of HiPerformer on eleven different datasets and compare it with various advanced models, validating its outstanding performance and effectiveness.   Our methods significantly improve feature representation capabilities, opening new avenues for related research. Despite its excellent performance, there are still some limitations, such as its IRMLP structure, which expands the dimensions by four times, resulting in a large overall number of parameters.  Future work should focus on optimizing the structure to reduce the parameter size and improve computational efficiency.

% or


% \appendices

% % you can choose not to have a title for an appendix
% % if you want by leaving the argument blank
% \section{}

% \section*{Acknowledgment}


\ifCLASSOPTIONcaptionsoff
  \newpage
\fi



%\bibliographystyle{IEEEtran}
% \input{HiPerformer.bbl}
\bibliographystyle{ieeetr}
% \bibliographystyle{unsrt}
% 要引用的文献

%%%%% CLEAR DOUBLE PAGE!
\newpage{\pagestyle{empty}\cleardoublepage}
% \cite{Chen2019Machine}
% \cite{Kushal2018Machine}

% \bibliography{ref}


% \end{IEEEbiography}


% % insert where needed to balance the two columns on the last page with
% %\newpage



%\vfill


\end{document}